\documentclass[sigconf]{acmart}

\AtBeginDocument{%
  \providecommand\BibTeX{{%
    \normalfont B\kern-0.5em{\scshape i\kern-0.25em b}\kern-0.8em\TeX}}}

\copyrightyear{2020} 
\acmYear{2020} 
\setcopyright{acmlicensed}\acmConference[WebSci '20]{12th ACM Conference on Web Science}{July 6--10, 2020}{Southampton, United Kingdom}
\acmBooktitle{12th ACM Conference on Web Science (WebSci '20), July 6--10, 2020, Southampton, United Kingdom}
\acmPrice{15.00}
\acmDOI{10.1145/3394231.3397905}
\acmISBN{978-1-4503-7989-2/20/07}

\usepackage{caption}
\usepackage{subcaption}
\usepackage{xspace}

\usepackage{tikz}
\usetikzlibrary{calc}
\usepackage{mwe}

\newcommand{\pb}[1]{\vspace{0.75ex}\noindent{\bf \em #1}\hspace*{.3em}}

\newcommand{\customxy}[3]{
    \begin{tikzpicture}
      \node[inner sep=10pt] (A) {\includegraphics[width=\linewidth]{#1}};
      \node[black] (B) at ($(A.south)!.05!(A.north)$) {#2};
      \node[black,rotate=90] (C) at ($(A.west)!.05!(A.east)$) {#3};
    \end{tikzpicture}
}

\def\etal{\emph{et al.}\xspace}

\begin{document}

\title{Analyzing Temporal Relationships between Trending Terms on Twitter and Urban Dictionary Activity}

%

\author{Steven R. Wilson$^1$, Walid Magdy$^{1,2}$, Barbara McGillivray$^{2,3}$, and Gareth Tyson$^{2,4}$}
\email{steven.wilson@ed.ac.uk, wmagdy@inf.ed.ac.uk,  bmcgillivray@turing.ac.uk, g.tyson@qmul.ac.uk}
\affiliation{%
  \institution{$^1$The University of Edinburgh, Edinburgh, UK}
  \institution{$^2$The Alan Turing Institute, London, UK}
  \institution{$^3$University of Cambridge, Cambridge, UK}
  \institution{$^4$Queen Mary University of London, London, UK}
}

\renewcommand{\shortauthors}{Wilson, Magdy, McGillivray, and Tyson}

\begin{abstract}
As an online, crowd-sourced, open English-language slang dictionary, the \emph{Urban Dictionary} platform contains a wealth of opinions, jokes, and definitions of terms, phrases, acronyms, and more. However, it is unclear exactly how activity on this platform relates to larger conversations happening elsewhere on the web, such as discussions on larger, more popular social media platforms. In this research, we study the temporal activity trends on Urban Dictionary and provide the first analysis of how this activity relates to content being discussed on a major social network: Twitter. By collecting the whole of Urban Dictionary, as well as a large sample of tweets over seven years, we explore the connections between the words and phrases that are defined and searched for on Urban Dictionary and the content that is talked about on Twitter. Through a series of cross-correlation calculations, we identify cases in which Urban Dictionary activity closely reflects the larger conversation happening on Twitter. Then, we analyze the types of terms that have a stronger connection to discussions on Twitter, finding that Urban Dictionary activity that is positively correlated with Twitter is centered around terms related to memes, popular public figures, and offline events. Finally, We explore the relationship between periods of time when terms are trending on Twitter and the corresponding activity on Urban Dictionary, revealing that new definitions are more likely to be added to Urban Dictionary for terms that are currently trending on Twitter.
\end{abstract}


\maketitle

\begin{figure*}
\centering
\begin{subfigure}{.33\textwidth}
  \centering
  \includegraphics[width=.9\linewidth]{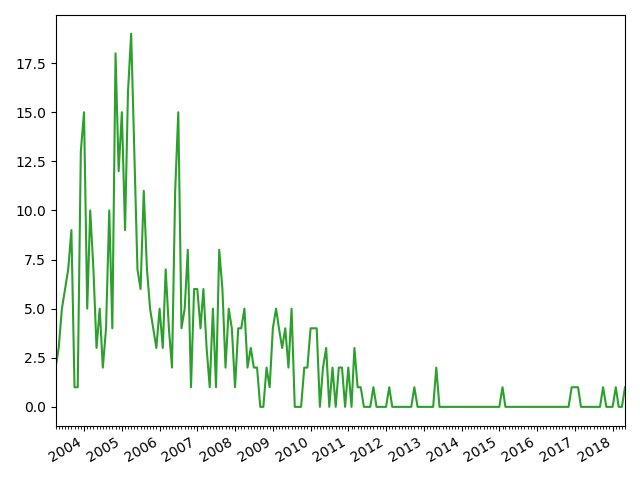}
  \caption{George W. Bush}
  \label{fig:sub-bush}
\end{subfigure}%
\begin{subfigure}{.33\textwidth}
  \centering
  \includegraphics[width=.9\linewidth]{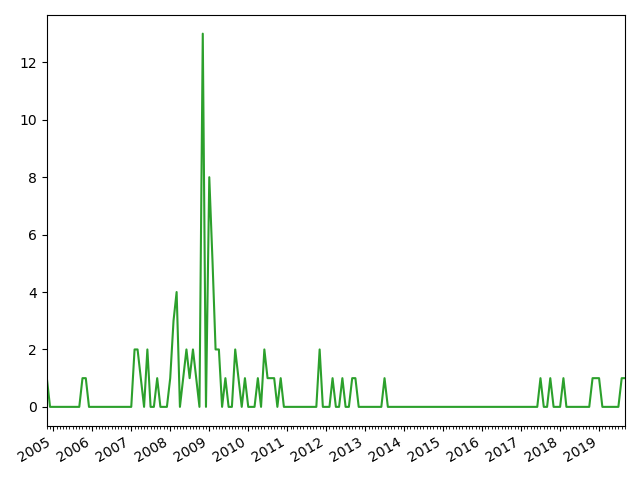}
  \caption{Barack Obama}
  \label{fig:sub-obama}
\end{subfigure}
\begin{subfigure}{.33\textwidth}
  \centering
  \includegraphics[width=.9\linewidth]{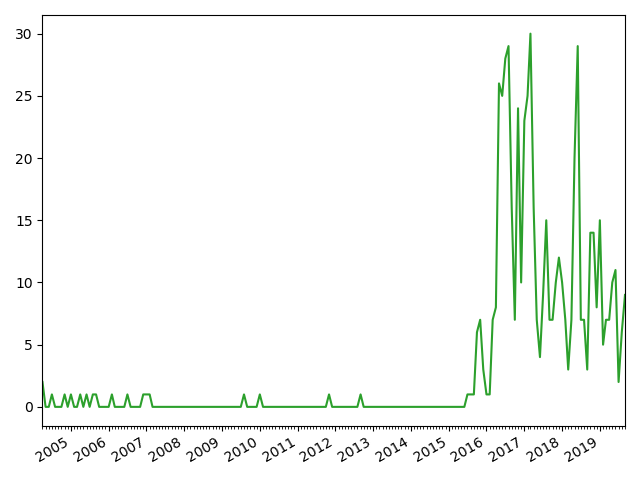}
  \caption{Donald Trump}
  \label{fig:sub-trump}
\end{subfigure}
\caption{Number of new Urban Dictionary definitions per month for the names of three recent U.S. Presidents.}
\label{fig:presidents}
\end{figure*}


\section{Introduction}

Online communities provide us with the means to study what people are interested in and talking about.
This includes political engagement~\cite{agarwal2019tweeting}, sports discussions~\cite{yu2015world} and general news~\cite{kwak2010twitter}.
However, these communities do not exist in isolation: the same users may visit multiple platforms, and information can propagate from one community to another.
For example, we regularly see this ecosystem effect when sharing memes~\cite{zannettou2018origins} and news media~\cite{zannettou2017web}.
Studying these kinds of connections can help us to learn more about how information moves across the web, and also can give us more insight into the way people are using various platforms.

In this study, we focus on the platform \emph{Urban Dictionary} (UD),\footnote{\url{https://www.urbandictionary.com/}} which is an online, crowdsourced dictionary for English slang and colloquial language.
Urban dictionary is known to be both complex and noisy, but also potentially invaluable in terms of its vantage of emerging slang terminology~\cite{nguyen2018emo}.
It serves as a mirror of parts of today's society, reflecting current trends and providing a perspective on the zeitgeist. For example, surges in definitions around U.S. Presidents George W. Bush (in office 2001-2009), Barack Obama (2009-2017) and Donald Trump (2017-Present) show how real-world events impact use of language online (Figure \ref{fig:presidents}). 

We posit that this connection to the zeitgeist may provide powerful insight into ongoing discussions, as well as offering a tool to better interpret online discourse.
However, to date, we lack the tools or computational studies that can measure the connection between UD and the kinds of conversations happening elsewhere on the web, e.g., Twitter. We are particularly interested in understanding how terminology may spread between platforms, and how UD influences with wider web-sphere. 

To overcome this deficiency, we present the first study to explore the relationship between UD and the use of terminology on a major social media platform, Twitter. We select Twitter due to its huge scale and ease of access to data. In this work, we specifically seek to answer the following research questions:
\begin{enumerate}
    \item Is activity on Urban Dictionary significantly correlated with discussions taking place on Twitter?
    \item If yes, for which terms does activity on these two platforms exhibit either a positive or negative temporal correlation? What are the characteristics of these terms?
    \item Is it more likely that new definitions are added to Urban Dictionary for a term if it is currently \textit{trending} on Twitter?
\end{enumerate}

To answer these questions, we collect minute-level data files containing tweets from a 1\% sample of all of Twitter between January 2012 and the end of September 2019, as well as a snapshot of the entirety of Urban Dictionary in October 2019. We use cross correlation analysis to explore the connections between activity on the two platforms, and we find that in some cases, UD activity \emph{does} reflect trends on Twitter, albeit with varying degrees of correlation and temporal lag. We categorize UD terms\footnote{Throughout the paper, we generically describe items that are defined in UD as ``terms'', while acknowledging that some of the headwords are actually multi-word expressions.} based on their association with Twitter, and find that positively correlated terms are more associated with political figures, memes, and historic events, while negatively correlated terms are more negative in sentiment, nonprofessional, and often have explicit themes. We also explore the relationship between \textit{trending} terms on Twitter and UD, finding that this tends to be strong in time periods connected to the creation of new definitions on UD.

\textit{We warn the reader that this paper contains offensive terms due to the nature of the data. It is necessary not to censor this content, so to offer a comprehensive description of material on Urban Dictionary.}

\section{Related Work}

\pb{Multi-Platform Analyses.}
There has been a recent surge in interest surrounding multi-platform influence. 
This includes understanding how news and links spread across websites~\cite{zannettou2017web}; how image content is copied between social media~\cite{zannettou2018origins}; and even how communities coordinate to impact other platforms~\cite{mariconti2019you}. 
These studies have shown that web and social platforms sit within a wider ecosystem with (poorly understood) influence over each other. We contribute to this understanding by inspecting how two particular platforms influence each other: UD and Twitter.

\pb{Evolution of Language \& UD.}
People have been studying the evolution of languages for hundreds of years~\cite{hamilton2016cultural}. This includes changes in word meanings~\cite{mitra2014s}, as well as how words are used~\cite{maity2016wassup,maity2016out}. Social media, however, has provided the first opportunity to get real-world insight into day-to-day changes in language \cite{shoemark-etal-2019-room}. 
We posit that UD better allows us to understand this evolution ``on the ground''.
There have been a small set of recent studies of UD. Smith \etal~\cite{smith2011urban} performed a qualitative analysis of how UD has effected and influenced both access to and formulation of the lexis.
Smith~\cite{smith2011urban}
performed a qualitative study, focusing on the word ``meep'', and exploring how UD might free language from prescriptive language ideologies. Wilson \etal~\cite{lrec2020} used UD as a training corpus for neural-network based word embeddings, finding that these embeddings were competitive with other popular pre-trained word embeddings models across a range of tasks including sentiment analysis and sarcasm detection. Closest to our work is that by Nguyen \etal~\cite{nguyen2018emo}, who performed a \emph{quantitative} study of terminology indexed on UD. They offer a statistical analysis of UD's content, showing for example a high presence of opinion-focused entries.

Our work differs in that we specifically look at how UD may influence other platforms. Furthermore, we focus on understanding ``activity log'' data, which was not inspected in these prior studies. 

\section{Methodology \& Data}
\label{sec:methodology}
We start by outlining our data collection methodology, as well as how we control for missing data.

\subsection{Urban Dictionary}

\begin{figure*}[tb]
\includegraphics[width=0.9\textwidth]{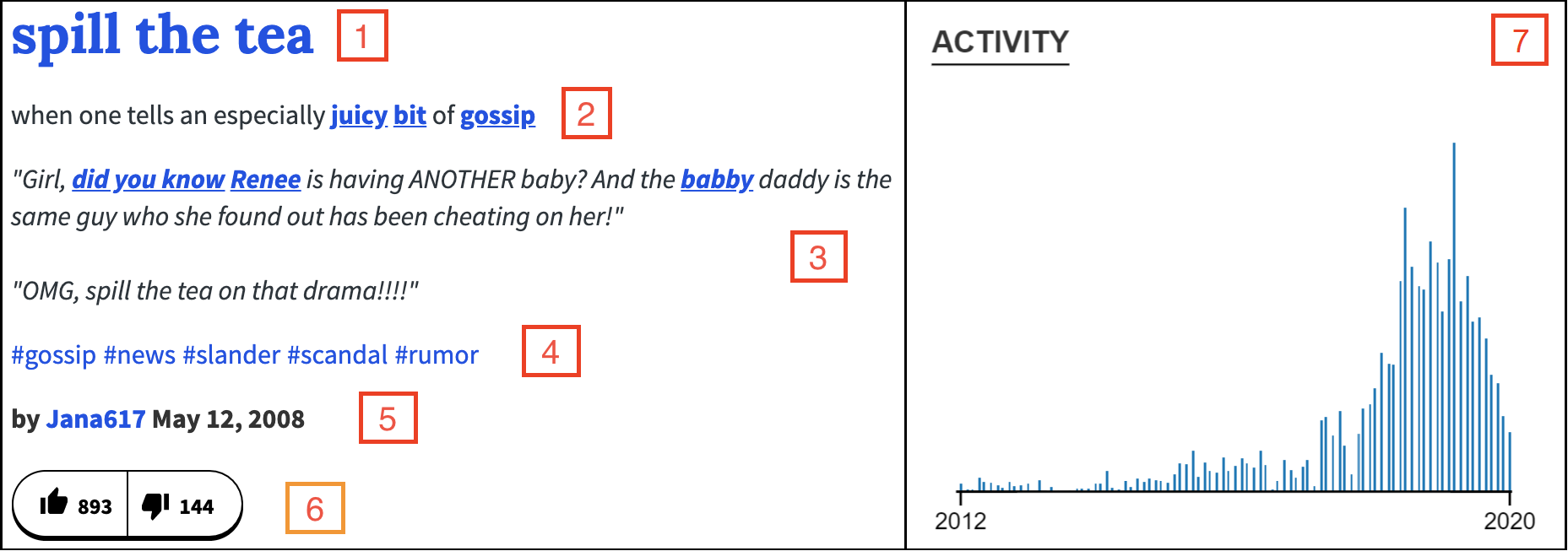}
\centering
\caption{Example entry on Urban Dictionary, including the head word (1), definition (2), usage examples (3), tags (4), user and date (5), upvote and downvote counts (6), and activity graph (7). Words and phrases in color that are also bold and underlined indicate links to other entries on Urban Dictionary.}
\label{fig:example}
\end{figure*}

Urban Dictionary is an online, crowd-sourced dictionary for (mostly)\footnote{Terms from other languages like ``hombre'' are defined, but definitions and examples describe code-switched usage of these terms within English speaking contexts.} English-language terms containing definitions that are not typically captured by traditional dictionaries. In the best cases, users provide meaningful definitions for new and emerging language, while in reality, many entries are a mix of honest definitions (``Stan: a crazy or obsessed fan''), jokes (``Shoes: houses for your feet''), personal messages (``Sam: a really kind and caring person''), and inappropriate or offensive language \cite{nguyen_ud}. 
Each entry, uploaded by a single user, contains a term, its definition, examples, and tags (Figure \ref{fig:example}). Further, those who view the entry have the opportunity to provide other definitions to the entry and/or also provide a vote in the form of a ``thumbs-up'' or a ``thumbs-down''. These votes are recorded and used to rank the possible definitions for a given term when it is looked up in Urban Dictionary. Entries in the Urban Dictionary can be for a singular word, a phrase (e.g., ``spill the tea'', Figure~\ref{fig:example}), or an abbreviation (e.g., ``brb'' and ``FYI'').

\begin{figure}[tb]
\centering
\includegraphics[width=\columnwidth]{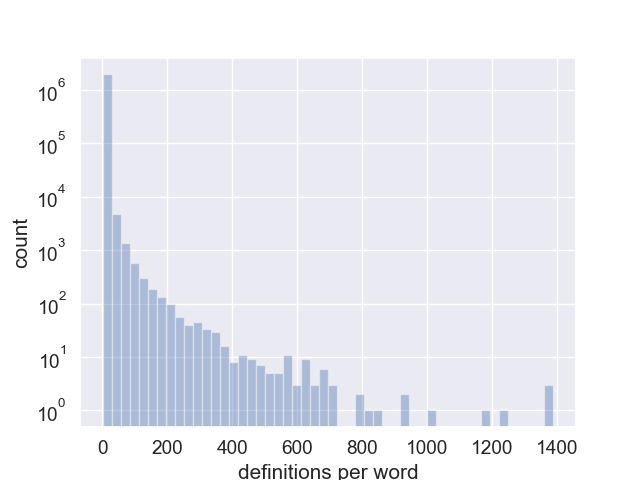}
\caption{Number of definitions per term defined on Urban Dictionary (log scale) \cite{lrec2020}.}
\label{fig:defperterm}
\end{figure}

\begin{figure}[tb]
\centering
\includegraphics[width=\columnwidth]{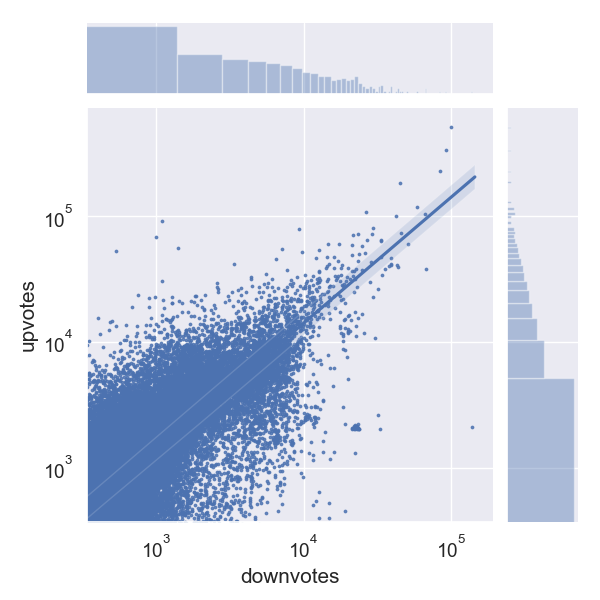}
\caption{Counts of upvotes and downvotes per entry on Urban Dictionary, with histograms (log scale) \cite{lrec2020}.}
\label{fig:votes}
\end{figure}

For every entry in Urban Dictionary, we crawl and store all of the aforementioned information, resulting in a total of approximately 2 million unique defined terms with an average of 1.8 definitions per term. The full histogram of the number of definitions per term is presented in Figure \ref{fig:defperterm}. This data collection includes an up-to-date version of Urban Dictionary as of October 16, 2019. In order to get a high-level understanding of the data, we also 
plot the upvotes and downvotes assigned to the full set of definitions in Figure \ref{fig:votes}. We note similar skewness in these figures as was reported in an earlier analysis of Urban Dictionary data \cite{nguyen_ud}.

We also scrape all “activity” statistics, which reflect user interest in these terms measured on month-to-month basis.
This is shown on the right hand side of Figure \ref{fig:example} (\#7) and represents the number of page clicks a definition has received.
We collect this information for all terms from January 2012 onward, since this is the earliest month for which this data is available across the site. As opposed to the temporal signal provided by the UD definitions, these activity statistics provide a more continuous gauge of overall interest in terms over time from a consumer perspective. These activity logs represent the number of visits to each word page over time. These are normalized, preventing us from known the scale of accesses. Instead, we can only see the trend. Note that these activity logs only cover 21.8\% of all terms, as less popular terms are not accompanied by the activity log.

\subsection{Twitter}\label{sec:Twitter}

We gather historical Twitter data from \url{archive.org},\footnote{\url{https://archive.org/download/archiveteam-json-twitterstream}} covering the same period as the Urban Dictionary activity statistics (i.e., starting in January 2012). 
This covers multiple terabytes of Twitter data, gathered using the 1\% “sprinkle” sample of the Twitter streaming API. Since UD is an English-language resource, we apply the pre-trained \texttt{fasttext} language classifier \cite{joulin2016fasttext} to all of the tweets, and only search for UD terms within the tweets that identified as being written in English. This is particularly important as UD contains a handful of terms, intended to be English slang or acronyms, that share surface forms with tokens in other languages (e.g., the Indonesian word ``nih'' will be confused with the UD term defined as an acronym for ``Not Invented Here'' or ``National Institute of Health''), leading to false positives. Further, we exclude words that are less than three characters long (the letters of the English alphabet have their own definitions on Urban Dictionary) or those that are included in a stopword list\footnote{English stopword list retrieved from \url{https://www.academia.edu/7221849/}}, leaving us with a set of 1,560,780 words and phrases to search for in each tweet. 

\subsection{Searching Twitter for UD terms}\label{sec:udsearch}

We check for all UD terms in each tweet using the Aho-Corasick algorithm \cite{aho1975efficient},\footnote{We use the implementation provided in the \hyperlink{https://pyahocorasick.readthedocs.io/en/latest/}{\texttt{pyahocorasick}} Python package.} which provides the locations of all substrings that match those in the input list to search for. We consider a term to be matched only if the characters before and after the substring match are both non-alphanumeric and if the string is not preceded with an $@$, indicating that the string is part of a handle (i.e., a username). We cannot first apply tokenization to the tweets, because some UD terms contain multiple tokens (e.g., ``falling in love'') or special characters like punctuation (``thebomb.com''), and so tokenization and other most text pre-processing steps would only make it more difficult to detect these terms. Therefore, we operate directly on the raw text of the tweets. The resulting total counts are then aggregated at the day-level, and the daily totals are then averaged across each month so that the length of a given month does not disproportionately affect its total count. 

\subsection{Missing Data.}
While our dataset represents a majority of the time period being studied, some segments of the Twitter data are missing for all terms.
We assume this was due to issues within the \url{archive.org} data collection.
To correct this, we check for any missing data at the minute-level and record the total number of minutes for which we have data each month. We define $O_m({M})$ as the \textit{observed} minute count for the month $M$ in a particular year. We then compute a correction for each month as:
$$
C(M) = \frac{E_m(M)}{O_m(M)}
$$
where $E_m(M)$ is the expected or actual number of minutes with month $M$. We estimate the number of minutes within a month as $60 \times 24 \times n_{days}$ where $n_{days}$ is the number of days during that month and year, taking leap years into account. We then take the total activity count $a(M)$ for each term found in month $M$ and multiply it by $C(M)$, rounding to the nearest whole integer, labeling this quantity, the corrected count for this month and year, as $\hat{a}(M)$. The average correction score across all months was 1.06, indicating that only a small number of total minutes were typically missing for a given month. In some instances, however, data is missing at the day level. For months missing more than 14 days of data,\footnote{These months were January 2014, January-March 2015, and May 2018.} we impute the counts of each term for that month by inserting the average of the (corrected) counts from the previous and following months.

\section{Cross-Platform Dynamics} 

We next proceed to explore key trends both within UD, as well as Twitter. We start by explaining how we selected key terms shared between both datasets, before proceeding to explore how these two platforms influence each other.

\subsection{Term Selection}\label{sec:term_selection}

Rather than examine every term in Urban Dictionary, we focus our study on the subset of terms that provide us with enough data to explore  interesting trends across our two platforms of interest. We consider all terms that:
\begin{enumerate}

\item have been defined on UD;
\item appear in our Twitter data sample at least 10,000 times over the course of nearly eight years of data; 
\item have recorded activity logs on UD that share at least 12 complete months of overlap with the available Twitter data.
\end{enumerate}

\begin{figure}
    \centering
    \customxy{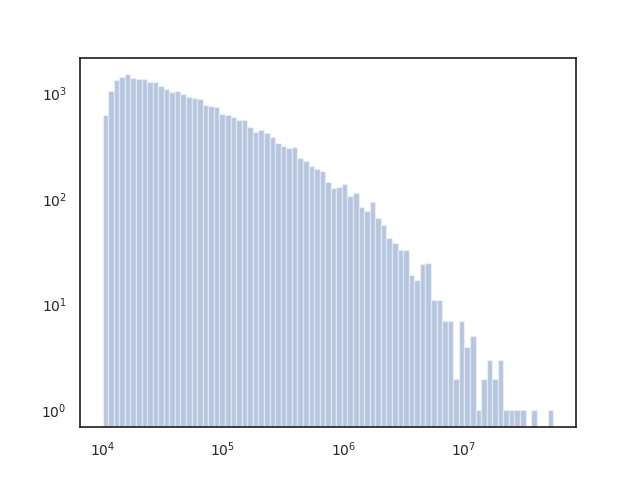}{UD term bin}{occurrences in Twitter sample}
    \caption{Histogram of total occurrences of selected (see section \ref{sec:term_selection}) UD terms in entire Twitter sample (log scale).}
    \label{fig:twitter_histogram}
\end{figure}

After applying these filtering steps, we are left with 31,803 terms, which appear in Twitter a total of 5,969,621,745 ($\approx$ 6 billion) times. The distribution of the total number of times that each of these terms appears in Twitter is presented in Figure \ref{fig:twitter_histogram}. Most of the terms appear between 10,000 (our minimum threshold value) and 1 million times, with a few appearing tens of millions of times through the time period we examine. Some of the most common UD terms on Twitter include ``lol'' (31 million occurrences), ``love'' (29 million),  ``twitter'' (17 million), ''retweet'' (16 million), and ``god'' (16 million). Interestingly, ``love'' and ``god'' are also two of the words that have previously been identified as having the largest number of distinct definitions on UD \cite{nguyen2018emo}. We spend the rest of the section exploring how these two time series datasets influence each other. 

\subsection{Who influences whom? Twitter or UD?}
\label{sec:cc}

We start by exploring how the use of terms with UD and Twitter correlate over time. 
Our goal is to understand if terms are introduced on Twitter and then spread to UD, or vice versa.
Specifically, measuring the cross-correlation between the two time series allows us to capture the relationships between the two sequences, as well as providing a measure of the time offset at which the two sequences are most highly correlated. Since the Twitter and UD data have differing units of measurement, we first normalize each month in time series, $S$, according to: 
$$
n(M,S) = \frac{\hat{a}(M) - \mu_S}{\sigma_S}
$$
where $\mu_S$ and $\sigma_S$ are the mean and standard deviation of the series $S$, respectively, and $\hat{a}(M)$ is the corrected activity value as computed in section \ref{sec:Twitter}, or the raw activity value in the case of UD. Then, define the series of all normalized values $n(M,S)$ for a given word as $S_w$, and let $U_w$ and $T_w$ represent the time series activity of term $w$ for UD and Twitter data, respectively. 

We can then measure the zero-normalized cross correlation as: 
$$
R_w(k,U,T) = \sum_{M\in X(U_w,T_w)} (n(M+k,U_w) \times \overline{ n(M,T_w) } )
$$
where $X(U_w,T_w)$ represents the longest overlapping period of time for which $U_w$ and $T_w$ are defined and $k$ represents a number of months. Call the time lag resulting in the most extreme positive or negative correlation $t_w=\operatorname{argmax}_k |R_w(k,U,T)|$ for $k\in[k_{min}~..~k_{max}]$. 

In order to split the terms based on those with a positive, negative, or no correlation, we identify the terms for which the difference between $R_w(t,U,T)$ and 0 is statistically significant with a value of $\alpha=0.01$,
correcting for multiple hypothesis testing using the Benjamini–Hochberg procedure \cite{benjamini1995controlling}\footnote{We use the implementation provided in the \hyperlink{https://www.statsmodels.org/dev/generated/statsmodels.stats.multitest.multipletests.html}{\texttt{statsmodels}} Python package.} to control the false discovery rate. When we find that we have sufficient evidence to reject our null hypothesis, $H_0: R_w(t,U,T) = 0$, we report that a term exhibits either a positive (if $R_w(t,U,T)>0$) or negative (otherwise) correlation between UD and Twitter activity with the defined $\alpha$ value. 

\begin{figure}
    \centering
    \begin{subfigure}{.45\textwidth}
        \centering
        \includegraphics[width=\linewidth]{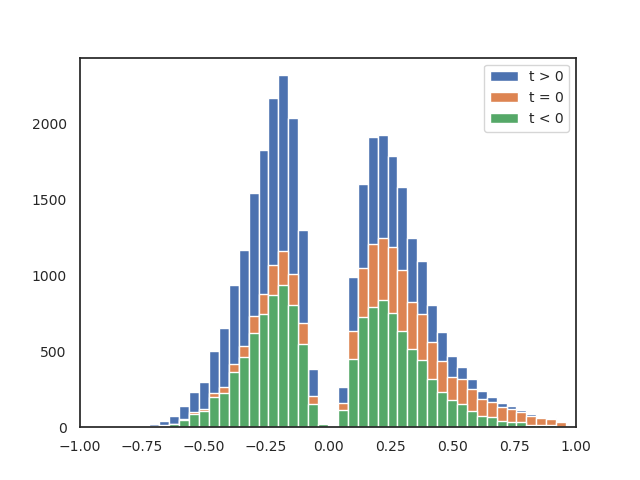}
        \caption{All terms.}
        \label{sub:before}
    \end{subfigure}%
    
    \begin{subfigure}{.45\textwidth}
        \centering
        \includegraphics[width=\linewidth]{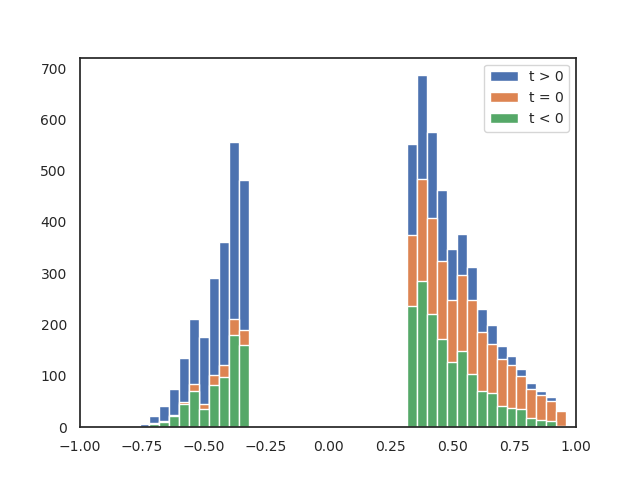}
        \caption{Only significantly correlated terms.}
        \label{sub:after}
    \end{subfigure}%
    
    \caption{Stacked histograms of most extreme temporal correlations between UD and Twitter activity trends considering possible time lags of $t=[-3~..~3]$ months before and after removal of cases where $H_0$ was rejected.}
    \label{fig:correlation_density}
\end{figure}

\begin{table}[]
\begin{tabular}{rcc|rcc}
\multicolumn{3}{c}{\textbf{Positive correlation}} & \multicolumn{3}{c}{\textbf{Negative correlation}} \\ \hline
term & corr & t & term & corr. & t \\ \hline
alex from target    & 1.000 & 0  & goth & -0.778 & 1 \\ 
number neighbor     & 1.000 & 0  & naruto & -0.721 & 1 \\
harlem shake        & 0.997 & 0  & mole & -0.720 & 3 \\
omarosa             & 0.993 & 0  & troll & -0.717 & 1 \\
pokemon go          & 0.991 & 0  & squirt & -0.699 & 2 \\
balsa               & 0.990 & 3  & as*hat & -0.698 & 0 \\
united airlines     & 0.989 & 0  & f*ck me & -0.691 & 3 \\
alternative facts   & 0.989 & 0  & pornography & -0.685 & 2 \\
franken             & 0.978 & 0  & f*cked & -0.676 & 3 \\
scaramucci          & 0.978 & -1 & hai & -0.676 & 3 \\
ebola               & 0.977 & -1 & p*ssy & -0.676 & 2 \\
lochte              & 0.977 & 0  & fisting & -0.675 & -3 \\
hurricane irma      & 0.975 & 0  & balls deep & -0.675 & 1 \\
kokobop             & 0.974 & 0  & fanboy & -0.674 & 3 \\
paris agreement     & 0.973 & -1 & squirting & -0.670 & 2 \\\hline
\end{tabular}
\caption{Examples of terms with strong positive and negative correlations between Twitter and UD activity trends, along with the value of $t$ (in $[-3~..~3]$) for which this correlation was measured.}
\label{tab:correlations}
\end{table}

We next proceed to discuss our results. The distribution of $R_w(t,U,T)$ for all values of $w$ is presented in Figure \ref{sub:before}, and the distribution for which the values are statistically significant in their difference from 0 is presented in Figure \ref{sub:after}. 
For context, the final value of $t_w$ tells us that the highest correlation for term $w$ occurs when $U_w$ is shifted by an offset of $t_w$ months. So, when $t_w$ is negative, we can say that the Twitter activity seems to lag \emph{behind} the UD activity, and when $t_w$ is positive, the \emph{opposite} is true. When $t_w=0$, the two time series seem to be most highly correlated with one another with no lag.

\begin{table}[]
\begin{tabular}{rc|rc}
\multicolumn{2}{c}{\textbf{Positive correlation}} & \multicolumn{2}{c}{\textbf{Negative correlation}} \\ \hline
tag & PMI & tag & PMI \\ \hline
\#rap & 0.843 & \#f*ckboy & 0.587 \\
\#politics & 0.771 & \#sensitive  & 0.579 \\
\#b*tches & 0.664 & \#big d*ck   & 0.527 \\
\#meme & 0.639 & \#pathetic    & 0.523 \\
\#omg & 0.563 & \#cheater	    & 0.477 \\
\#internet & 0.559 & \#personality	& 0.469 \\
\#ghetto & 0.521 & \#creative	& 0.452 \\
\#school & 0.515 & \#bestfriend	& 0.445 \\
\#poser & 0.485 & \#america	    & 0.436 \\
\#wtf & 0.467 & \#pleasure	& 0.430 \\\hline
\end{tabular}
\caption{Examples of UD tags applied to terms with with significant positive and negative correlations between Twitter and UD activity trends.}
\label{tab:pmi}
\end{table}

\begin{figure*}
    \centering
    \begin{subfigure}{.33\textwidth}
        \centering
        \includegraphics[width=.9\linewidth]{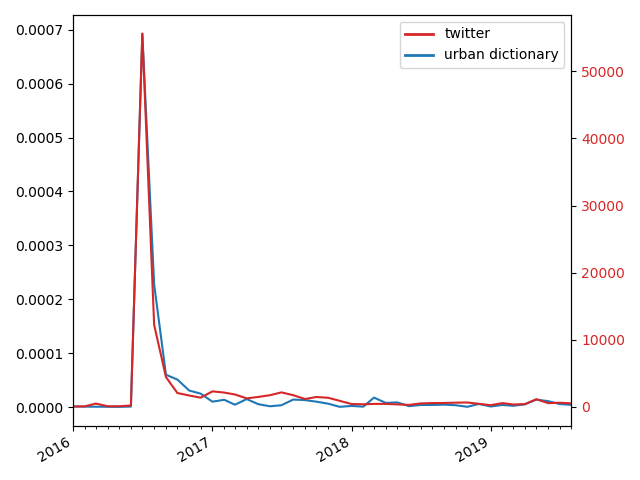}
        \caption{Pok\'{e}mon Go: positive correlation ($0.99$)}
    \end{subfigure}%
    \begin{subfigure}{.33\textwidth}
        \centering
        \includegraphics[width=.9\linewidth]{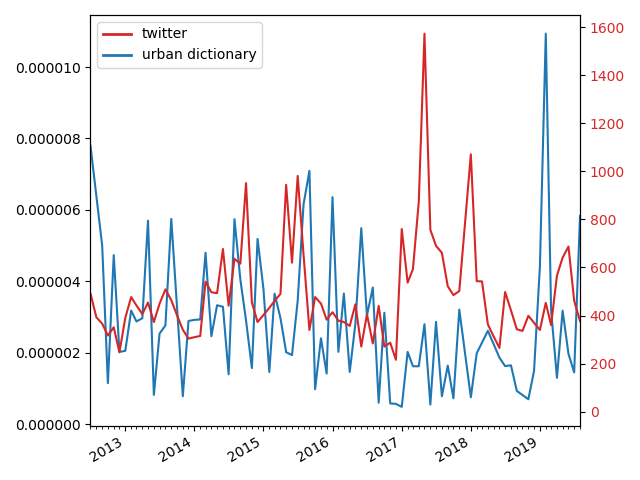}
        \caption{Guacamole: no sig. correlation ($-0.11$)}
    \end{subfigure}
    \begin{subfigure}{.33\textwidth}
        \centering
        \includegraphics[width=.9\linewidth]{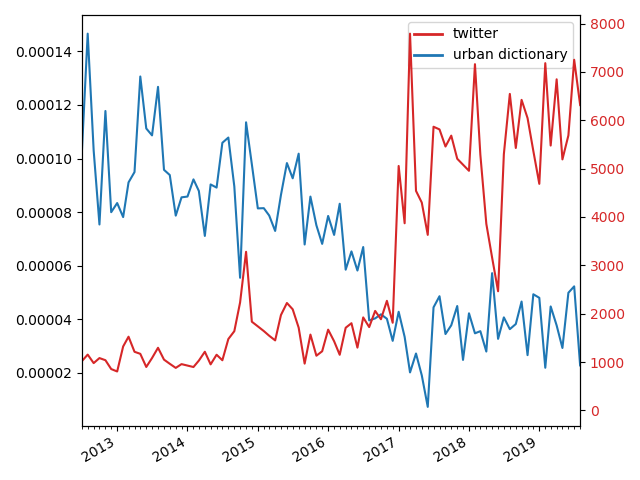}
        \caption{Naruto: negative correlation ($-0.72$)}
    \end{subfigure}
    \caption{Month-level plots of three example terms with correlated and uncorrelated activity on Urban Dictionary and Twitter over time. Pok\'{e}mon Go is an augmented reality mobile phone game that was initially released on July 6, 2016, exhibiting highly focused attention. The cross-platform interest in the term ``guacamole'' shows no consistent patterns over time. For the Japanese manga Naruto, increases in Twitter discussion in early 2017 align with the end of a ten year television series, while activity on Urban Dictionary begins to drop-off around the same time.}
    \label{fig:corr_example}
\end{figure*}

Figures \ref{sub:before} and \ref{sub:after} show that there are, indeed, noticeable correlations between the use of terminology on Twitter and its definition in UD. 
It is marginally more typical for terms to emerge on Twitter before UD, rather than vice versa.
Overall, we identify 4,917 terms for which Urban Dictionary and Twitter activity is correlated. 
To provide context, Figure~\ref{fig:corr_example} provides prominent examples of three terms that have positive, negative and no correlations. We see noticeable differences with viral terms like ``Pok\'{e}mon'' highly correlated. Further examples of these terms are presented in Table \ref{tab:correlations}. 
For instance, we see that for certain well known and longstanding terms (e.g., ``goth'' and ''f*cked''), Twitter lags behind UD, but for other more emergent terms and memes (e.g., ``alex from target'', ``pokemon go'' and ``harlem shake'') Twitter is ahead of UD. This suggests that terminology usage requires a critical mass, before warranting inclusion on UD. We also see cases where sudden events (e.g., ``hurricane irma'') rapidly emerge on Twitter, before later being added to UD.
Briefly, we also examine the number of likes and dislikes given to definitions of these words on UD, finding no major differences from the overall distribution (originally presented in Figure \ref{fig:votes}).

\subsection{What themes are defined and discussed?}

We next inspect which themes are covered within these terms. To achieve this, we use hashtags associated with each term as a proxy (each definition can be accompanied by tags).
First, we take the set of tags given by UD users to each of the terms and compute the point-wise mutual information (PMI) between the occurrence of the tag and one of three categories \cite{manning2008introduction}. Specifically, we categorise terms based on whether or not their usage is correlated on Twitter with UD (as defined in Section \ref{sec:cc}). For simplicity, we group each term into: positive correlation, negative correlation, or no correlation. PMI is computed as
$$
PMI(x,y) = \operatorname{log}\frac{\operatorname{p}(x,y)}{\operatorname{x}\operatorname{y}}
$$
where, in our case, $x$ is a variable representing the event that a tag is attached to a term and $y$ represents the event that a term belongs to the set of either positively correlated or negatively correlated time series. The joint probability $\operatorname{p}(x,y)$ represents the likelihood that a specific tag has been assigned to a term that also belongs to a category: positive, negative, or no correlation, and we can compute a PMI score for each tag for each set. 
Note that we consider the full set of tags, including those assigned to the ``not correlated'' group, when computing the observed probabilities of tags or categories occurring, though we are only interested in computing the final PMI scores for the positively correlated and negatively correlated categories. 

\begin{table}[]
\begin{tabular}{r|l|l|l|l}
                     & t < 0 & t = 0 & t > 0 & \textbf{all} \\ \hline
positive correlation & 75.8\% & 63.9\% & 72.5\% & \textbf{70.0\%} \\ 
no significant correlation & 79.01\% & 79.4\% & 81.6\% & \textbf{80.2\%} \\
negative correlation & 94.6\% & 94.2\% & 89.7\% & \textbf{93.1\% }\\ \hline
\textbf{all} & \textbf{82.1\%} &\textbf{ 77.4\%} & \textbf{80.9\%} & \textbf{79.8\%} \\ 
\end{tabular}
\caption{Percentage of terms with definitions in Wiktionary.}
\label{tab:wiktionary}
\end{table}

The tags with the highest PMI scores for the positive and negative correlation groups are presented in Table \ref{tab:pmi}. 
We are particularly curious to understand if these terms with significant correlations are nonstandard English words, multi-word expressions, or proper nouns. To explore this, we compute the percentage of each group that has been defined in the English section of the online resource Wiktionary.\footnote{\url{https://en.wiktionary.org/}} 
Table \ref{tab:wiktionary} shows the proportion of terms that are defined in Wiktionary for each cross-section of data based on level of correlation and value of $t$. Interestingly, we observe that the greatest fraction of terms that are \textit{undefined} in Wiktionary come from the ``positive correlation'' group (note the lower overall fraction of terms with definitions for this group, the first row in Table \ref{tab:wiktionary}) indicating that words from this group are less likely to be standard English words.

\subsection{Are UD entries more likely for trending terms?} \label{sec:trends}

We conjecture that certain terms may experience rapid surges in popularity, and that these surges may correlate with new entries being added for terms on UD. Thus, we next explore if certain terms start to ``trend'' at points within our measurement period, both within Twitter and UD, and how likely it is that new entries are added to UD for terms that are currently trending. Previously proposed trending detection algorithms typically act in real time, relying only the use of data preceding the point of the trending period in order to detect trends as early as possible \cite{xie2016topicsketch}. Trend detection approaches may also involve the use of machine learning models that are trained to recognize examples of items that were known to go on to be considered trending \cite{chen2013latent}. However, these approaches depend on knowledge of ``ground truth'' for which terms eventually moved into a \textit{trending} period, meaning that a potentially unknown definition of \textit{trending} is being learned. Others approaches aim for personalization by incorporating user-level features such as the types of topics that a person is typically interested in \cite{fiaidhi2013developing}, which we do not make use of as we are searching for general periods of upward trending. Additionally, we do not consider burst detection methods \cite{kleinberg2003bursty} which can accurately identify abnormal spikes in usage, since we also wish to discover trends that experience a rapid initial increase in usage followed by long plateaus of high usage, e.g., for terms that were first introduced at some point in time yet remained popular after the initial increase in usage.

As we are able to analyze the entire period of interest \emph{post-hoc}, and we would like to apply criteria for trending detection that are general to both UD and Twitter, we opt for the following approach. Inspired by previous work in the earth science domain \cite{sharma2016trend}, we fit a piece-wise function across the entire time series. This allows us to quickly check for sections of rapid increases by analyzing the slope of this function at a given point in time.

\begin{figure}
    \centering
    \customxy{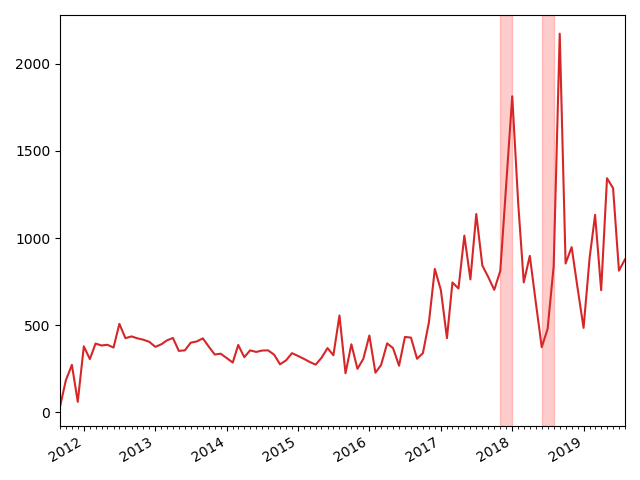}{year}{occurrences in Twitter sample}
    \caption{Twitter activity plot for the term ``Vibing'' along with detected \textit{trending} periods (shaded regions) using the proposed approach.}
    \label{fig:trending}
\end{figure}

To fit the piece-wise function, we first split the time series at all identified change points using the pruned exact linear time (PELT) change point detection algorithm \cite{killick2012optimal}\footnote{We use the implementation provided in the \hyperlink{http://dev.ipol.im/~truong/ruptures-docs/build/html/index.html}{\texttt{ruptures}} Python library.}. 
PELT is a dynamic programming approach used to efficiently find the best segmentation of a time series by minimizing a cost function defined in terms of the likelihood of the data in each segment. After running PELT on each time series, we then fit an ordinary least squares regression line to each segment of the data that lies between two change points, and inspect the slope of the line. If the \textit{slope} is greater than the threshold $\tau_m$, we mark this period of time as ``trending'' for this term. In our analyses, we set $\tau_m=\frac{\operatorname{max}(S)}{4}$ where $S$ represents all points in a time series. Figure \ref{fig:trending} shows an example of the results of this trending detection approach on our Twitter data sample.

\begin{table}[]
\begin{tabular}{p{2cm}cc}
 & \textbf{Twitter} & \textbf{UD} \\ \hline
$\operatorname{p}(d|u)$        & \textbf{0.105} & \textbf{0.113} \\
$\operatorname{p}(d|\neg u)$   & 0.077 & 0.104 \\ \hline
$\operatorname{p}(u|d)$        & \textbf{0.142} & \textbf{0.172} \\
$\operatorname{p}(u|\neg d)$   & 0.111 & 0.162 \\ \hline
\end{tabular}
\caption{Observed probabilities associated with the creation of new definitions on UD and trending periods on Twitter (column 1) and UD (column 2). Bold font denotes a statistically significant difference from the quantity directly below in the table using a two sample t-test and $\alpha=.001$. $d=1$ indicates that a term is defined in a given month, and $u=1$ indicates that the same term is \textit{trending} on either Twitter or UD during that month.}
\label{tab:probs}
\end{table}

We compute all trending periods for both the UD and Twitter time series for all terms, and compare these time periods to the dates during which new definitions are added to UD for a given term. Let the symbol $d$ represent a binary random variable that is \texttt{true} in the event that a new definition for a term is added during a given month, and \texttt{false} otherwise. Then, let $u$ be \texttt{true} if term the same term is \textit{trending} during month the same month. We estimate the conditional probabilities associated with various values of $d$ and $u$ in Table \ref{tab:probs}. We find that the probability of observing a new definition for a given term is statistically significantly more likely in a given month if activity centered around that term is \textit{trending} on either UD or Twitter. Further, when a term has received a new definition in a given month, it is also more likely that this term would be marked as \textit{trending} according to our trend detection algorithm for either the UD activity or the Twitter activity time series.

\section{Discussion}

Having completed our analyses, we return to our initial research questions and attempt to answer each given the evidence that we have gathered.

\pb{(1) Is any activity on Urban Dictionary significantly correlated with discussions taking place on Twitter?} In section \ref{sec:cc}, we computed the cross-correlations between the monthly Twitter and UD activity time series and found that, for a subset of terms of interest, there was a significant correlation between activity on the two platforms. While we are unable to make conclusions about the majority of terms that appear on UD and Twitter, we are able to identify those terms for which there exists either a positive or negative correlation. Overall, we find that there are more terms with a significant positive correlation, and that these correlations occur with a time lag of 0, suggesting that the activity that is happening on UD and Twitter is generally synchronized for these terms. These results confirm that UD itself does in fact reflect trends occurring elsewhere around the internet, and based on qualitative analyses, events taking place in the offline world.

\begin{figure}
    \centering
    \includegraphics[width=\linewidth]{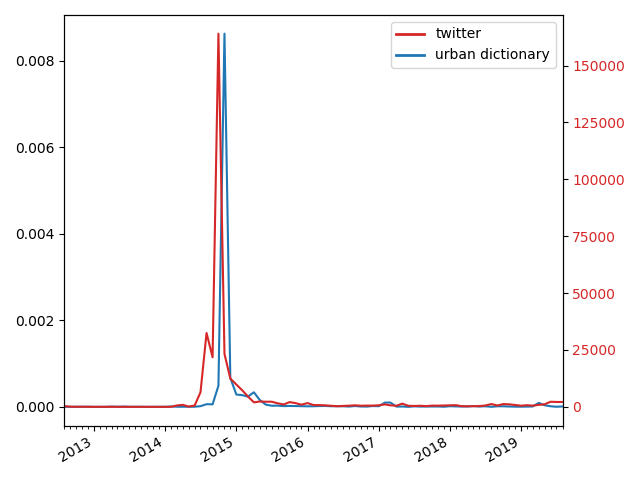}
    \caption{Twitter and UD activity for the term ``ebola''.}
    \label{fig:ebola}
\end{figure}

\pb{(2) If yes, for which terms does activity on these two platforms exhibit either a positive or negative temporal correlation? What are the characteristics of these terms?}
As we did find a link between the activity on the two platforms, we explored some of these terms and their attributes later in Section \ref{sec:cc}. We notice several major trends for the terms exhibiting positive correlations between Twitter and UD activity measurements. First, we see a theme of internet memes, exemplified by the terms ``alex from target'', ``harlem shake'', and ''number neighbor'', as well as tags such as \#meme and \#internet. 
Second, there are a myriad of terms related to political figures and large-scale events, such as ``omarosa'', ``scaramucci'', ``ebola'', ``hurricane irma'' and ``paris agreement'', as well as the tag \#politics. Since many of these terms are related to extremely specific events that took place in a single month or even a single day, the online activity observed on both platforms is often very acutely focused around the time of that event. For example, see the time series plot of the term ``ebola'' in Figure \ref{fig:ebola}. There is a single major spike in both time series in late 2014, roughly when the first case of the Ebola virus was confirmed in the United States during the 2014-2016 epidemic \cite{kaner2016understanding}. For the negatively correlated terms, we instead see a range of slang and risqué language. While further investigation is needed to fully understand why these terms exhibit a strong negative correlation between UD and Twitter activity, one possibility is that we may be tapping into a larger trend taking place on these platforms in which language that was once considered taboo and was relegated only to website like UD is now more well known and commonplace, appearing more on Twitter, making it less novel on UD. Either way, it is clear that these two platforms \emph{do} influence each other (either tacitly or directly).

\pb{(3) Is it more likely that new definitions are added to Urban Dictionary for a term if it is currently \textit{trending} on Twitter?}
In Section \ref{sec:trends} we define \textit{trending} for a given term on either platform. Given this definition and the data we have about the creation of new definitions for terms of UD, we calculate the likelihood of terms appearing both inside and outside of \textit{trending} intervals, finding it (statistically) significantly more likely to witness new definitions during \textit{trending} periods when considering both UD and Twitter time series. Additionally, we find that terms are more likely to be \textit{trending} during months for which new definitions have been added to UD. While capturing the causal relationships at play, if they exist, is left as future work, these results solidify the relationship that exists between the observed user behaviors on these two platforms centered around specific types of content.

\section{Conclusion}
We have presented the first analysis of the temporal relationships between online activity on the under studied platform Urban Dictionary and the broad conversations happenings on Twitter. We explored the relationships between periods of time when terms were \textit{trending} and corresponding activity on Urban Dictionary, such as the creation of new definitions, finding that new definitions are more likely to occur during these periods. Through a series of cross-correlation analyses, we identified cases in which Urban Dictionary activity most closely reflects the content being discussed on Twitter. By inspecting and characterizing the types of terms that have a stronger connection to discussions on Twitter, we found that Urban Dictionary activity that is positively correlated with Twitter mentions is centered around terms related to memes, popular public figures, and offline events. While this work represents an initial venture into the study of the links between these two platforms, we hope that it provides a foundation for future work exploring the web and its many components as a larger socio-technical system, searching for interactions between various online communities and their behaviors rather than studying each one in isolation.

\begin{acks}
This work was supported by The Alan Turing Institute under the EPSRC grants EP/N510129/1, and EP/S033564/1. We also acknowledge support via EP/T001569/1.
\end{acks}

\bibliographystyle{ACM-Reference-Format}
\bibliography{websci}



\end{document}